\documentclass[letterpaper, 10 pt, conference]{ieeeconf}  

\IEEEoverridecommandlockouts                              
\usepackage{graphicx}
\usepackage{subfigure}
\usepackage{amsmath}
\usepackage{algorithm}
\usepackage{algorithmic}
\usepackage{makecell}
\usepackage{color}
\usepackage{txfonts}
\title{\LARGE \bf
Identification and Avoidance of Static and Dynamic Obstacles on Point Cloud for UAVs Navigation
}

\author{Han Chen$^1$ and Peng Lu$^{1,2*}$ 
\thanks{$^1$The Adaptive Robotic Controls Lab (ArcLab), Hong Kong Polytechnic University, Hung Hom, Kowloon, Hong Kong, China.
{\tt\small stark.chen@connect.polyu.hk}}
\thanks{$^2$Department of Mechanical Engineering, The University of Hong Kong, Pokfulam, Hong Kong, China.
{\tt\small lupeng@hku.hk}}
\thanks{$^*$Corresponding author}}%

\begin{document}

\maketitle
\thispagestyle{empty}
\pagestyle{empty}

\begin{abstract}

Avoiding hybrid obstacles in unknown scenarios with an efficient flight strategy is a key challenge for unmanned aerial vehicle applications. In this paper, we introduce a technique to distinguish dynamic obstacles from static ones with only point cloud input. Then, a computationally efficient obstacle avoidance motion planning approach is proposed and it is in line with an improved relative velocity method. The approach is able to avoid both static obstacles and dynamic ones in the same framework. For static and dynamic obstacles, the collision check and motion constraints are different, and they are integrated into one framework efficiently.
In addition, we present several techniques to improve the algorithm performance and deal with the time gap between different submodules. The proposed approach is implemented to run onboard in real-time and validated extensively in simulation and hardware tests. Our average single step calculating time is less than 20 ms. 

\end{abstract}

\section{INTRODUCTION}

In unknown and chaotic environments, unmanned aerial vehicles (UAVs), especially quadcopters always face rapid unexpected changes, while moving obstacles pose a greater threat than static ones. To tackle this challenge, the trajectory planner for UAVs needs to constantly and quickly generate collision-free and feasible trajectories in different scenarios, and its response time needs to be as short as possible. In addition, the optimality of the motion strategies should also be considered to save the limited energy of quadrotors.

 Most existing frameworks that enable drones to generate collision-free trajectories in completely unknown environments only take into consideration stationary obstacles. However, as quadrotors often fly at low altitudes, they are faced with many moving obstacles such as vehicles and pedestrians on the ground. One primary solution to avoid collision is to raise flight altitude to fly above all the obstacles. This method is not feasible for some indoor applications, because the flight altitude is lower in narrow indoor space, and the drones are often requested to interact with humans as well. Another solution is to assume all detected obstacles as static. But this method cannot guarantee the safety of the trajectory \cite{c18}, considering measurement errors from the sensors and unmissable calculating time of the planner. Therefore, a more efficient and safer way to avoid moving obstacles is to predict and consider the obstacles’ position in advance based on the velocity, which can avoid detours or deadlock on some occasions.
	
 
   \begin{figure}[t]
      \centering
      \includegraphics[width=0.47\textwidth]{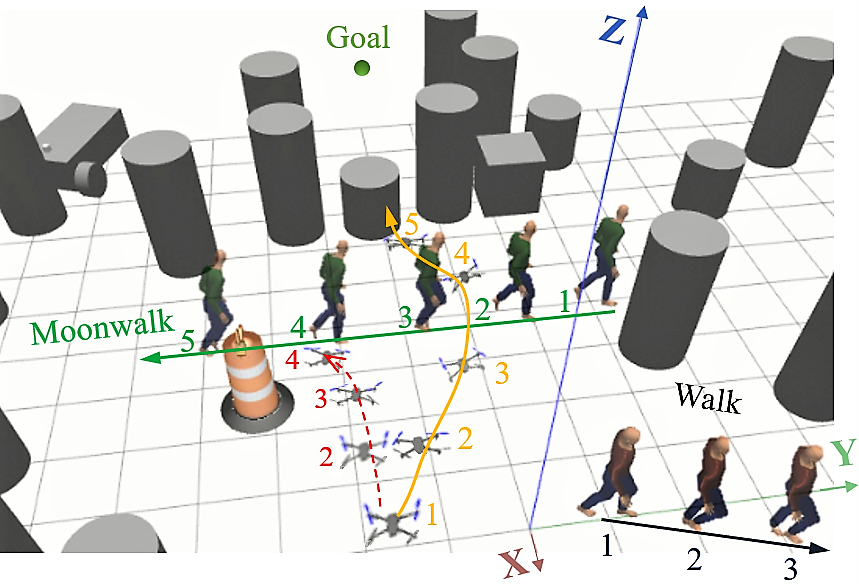}
      \caption{The composite picture of the simulation in Gazebo for the process that the drone avoids static and dynamic obstacles. 5 screenshots are used for composition and the cut time interval is fixed to 0.7 seconds. The line with an arrowhead shows the moving direction and the numbers mark the corresponding frame, the numbers increase by time. The yellow line is generated by the method in this paper, while the red line is by the original static method. The corresponding visualized data is shown in Fig. \ref{fig11}(c).}
      \label{fig1}
      \vspace{-0.5cm}
   \end{figure}
In this paper, we propose a framework to distinguish and avoid the dynamic and static obstacles, including a raw point cloud filter, a position and velocity estimator for moving  obstacles, and a motion planner to set the waypoint and generate motion primitives. Considering the fast flight and the rapid change of the environmental information, a safe and effective obstacle avoidance algorithm is needed to process up-to-date information from the drones during flight. Thus, the relative velocity \cite{c24} method is applied for static and dynamic obstacle collision check. It is combined with our former proposed heuristic angular search (HAS) \cite{c26} method to avoid both static and dynmaic obstacles efficiently. With point cloud input, a cluster algorithm is adopted to divide the point cloud into isolate obstacles, then the static and dynamic obstacles can be sorted by the displacement between two sensor frames. A depth camera is utilized to obtain the point cloud. To overcome its narrow field of view (FOV), the static points are memorized to prevent the drone from hitting obstacles that just moved outside the FOV. For moving obstacle avoidance, the information (position, velocity and size) of the velocity obstacle is used to choose a safe path and smaller acceleration for the drone. This is better than considering all the points as static obstacles.

In summary, the main contributions of the paper are as follows:

\begin{itemize}

\item  A novel computational efficient framework is proposed in this paper for UAVs navigation to avoid dynamic and static obstacles concurrently on point cloud with a near optimal strategy.
\item We introduce the \textbf{“spherical hypothesis”} to reduce errors when estimating the obstacle position and use a Kalman filter to make the estimation output more accurate and stable. The position, size, and velocity of the moving obstacles are estimated from the point cloud. 
\item Based on the HAS and relative velocity method, a \textbf{velocity planning algorithm} is designed to make the planner more effective and try to reduce the acceleration cost while guaranteeing safety.
\end{itemize}

In addition, we conduct a number of simulation and hardware tests to prove the proposed planner is effective. Furthermore, the computation time rarely increases (only up about 5 ms) compared to the method which does not take into account the velocity of obstacles \cite{c26}.
\section{RELATED WORK}

At present, there are many methods for path planning and obstacle avoidance in static environments. Although some researchers have published their safe planning framework for UAVs in an unknown static scenario, it is a more complex problem for a UAV with a single depth camera flying in an unknown environment with dynamic obstacles. Because the FOV is narrow and the pose of the camera is unstable during the flight. For hardware experiments, it is necessary to encode and use the information of detected obstacles in an efficient way to make the drone react agilely to moving obstacles in advance.

For identifying moving obstacles from the environment, most researchers employ the raw image from the camera and mark the corresponding pixels before measuring the depth. The semantic segmentation network with moving consistency check method based on images can distinguish the dynamic objects \cite{c20}. Also, a block-based motion estimation method to identify the moving obstacle is used in \cite{c21}, but the result is poor if the background is complex. If only human is considered as moving obstacle, the human face recognition technology can be applied \cite{c2}. However, the above-mentioned works do not estimate the obstacle velocity and position.
\cite{c23} proposes such an approach, which jointly estimates the camera position, stereo depth and object detections, and it can track the trajectories. But all the above image-based method is computation-expensive, not appropriate for the onboard micro computer. Based on the point cloud from lidar, it is also possible to estimate the moving obstacles' position and velocity \cite{c22}. This inspired our own approach in this paper. Event cameras can distinguish between static and dynamic objects and enable the drone to avoid the dynamic ones in a very short time \cite{c14}. However, the event camera is expensive for low-cost UAVs and not compatible to sense the static obstacles.

In terms of the avoidance of moving obstacles for navigation tasks, the majority of research works are based on the applications of ground vehicles. The forbidden velocity map \cite{c24} is designed to solve out all the forbidden 2D velocity vectors and they are represented as two separate areas in the map. The artificial potential field (APF) method can avoid the moving obstacles by considering their moving directions \cite{c3}-\cite{c15}. Also, the model predictive control (MPC) method is tested, but the time cost is too large for real-time flight. \cite{c7} proposes the probabilistic safety barrier certificates (PrSBC) to define the space of admissible control actions that are probabilistically safe, which is more compatible for multi-robot systems. \cite{c12} adopts the similar relative velocity concept with this paper, while \cite{c12} is in the 2D scenario and it samples from the velocity space rather than find the optimal solution.

For generating motion primitives and predicting the trajectory, one typical solution is to build a local occupancy grid map with the most recent perception data and generate a minimum-jerk trajectory through the waypoints \cite{c4}-\cite{c5}. A bi-level optimization was used in \cite{c6}-\cite{c8} to find the time constrains for motion optimization. Additionally, the motion primitives can be obtained by solving an optimization problem directly without searching for the waypoints \cite{c10}-\cite{c25}. Usually, a Bezier curve is used to express the predicted trajectory for its convex property. Another way is to directly generate motion primitives by sampling \cite{c13}, then select the most suitable group of motion primitives by an evaluation function as the output. One representative work is \cite{c9}, the drone can be precisely controlled to fly an intersecting trajectory with a falling ball and catch it. Except the sampling-based methods, it is difficult to significantly increase the calculation speed within their own framework.



\section{TECHNICAL APPROACH}
\vspace{-0.1cm}
\begin{table*}[h]
    \caption{definition of mathematical symbols}
\vspace{-0.6cm}
\label{table_1}
\begin{center}
\begin{tabular}{|p{1.2cm}<{\centering}|p{7.0cm}<{\centering}|p{1.2cm}<{\centering}|p{6.9cm}<{\centering}|}
\hline
$\overline{\quad},\ \overrightarrow{}$ & the line segment and vector between the two points & $Pcl_{t_1},Pcl_{t_2}$ & the former and latest frames of $Pcl_2$, at timestamp $t_1$ and $t_2$ \\
\hline
 $p_{ali},\ e_{ali}$ & the UAV position \& orientation aligned to $Pcl_2$ & $p_{t_1},\ p_{t_2}$ & position of the UAV at timestamp $t_1$ \& $t_2$\\
\hline
 $e_{t_1},\ e_{t_2}$ & orientation of the UAV at timestamp $t_1$ \& $t_2$ & $OT_{1}$, $OT_{2}$ & lists for point clusters and their centers in $Pcl_{t_1}$ and $Pcl_{t_2}$\\
\hline
 $C_{1}$, $C_{2}$ &lists for clusters centers in $OT_{1}$ and $OT_{2}$ & $ot_{1}$, $c_1$ & any point cluster in $OT_1$, and the corresponding center\\
\hline
 $len()$&function returns the length of the input list& $D(),\ V()$ & return the points coordinate variance and obstacle volume \\
 \hline
 $C_{M}(),C_{V}()$&return the mean and variance of the points color value &  $C_d$ & the list for storing the moving obstacle position $c_{2k}$\\
\hline
$R,\ H,\ Q$& observation noise covariance, observation, and error matrix & $P_{t}$ & posterior estimation error covariance\\
\hline
$\hat{c}_{2k}$, $\hat{v}_{2k}$ & filtered position and velocity of $ot_{2k}$ & $a_{t-1}$ & observed obstacle acceleration \\ 
 \hline
$p_{n},\ v_{n},\ a_{n}$& the position, velocity and acceleration of UAV for motion planning at current step &$abs()$ & returns the vector composed of the absolute values of all the input vector elements\\
\hline
$ag()$ & returns the direction angle of the given vector & $rd()$ & returns the closest integer of the input number\\
\hline
$\alpha_0$ & the direction angle of the original relative velocity& $\hat C_d$& the list storing all $\hat c_{2k}$ computed by the Kalman filter\\
\hline
$r_{obs}$& the size of the moving obstacle  & $r_{safe}$ & safety radius, a pre-assigned constant\\
\hline
 $w_{p(n)}$ & the point where the planned trajectory try to end & $argmin()$&returns the label of the minimum element in a vector\\
\hline
$A_t$ & matrix to solve $s_p$ & $v_{unit}$ & unit vector for the current search direction of HAS\\
\hline
$s_{p}(1)$, $s_{p}(2)$& magnitudes of velocity and relative velocity planned for the drone next step & $V_{g},\ v_g$ & the list of all the feasible UAV velocity magnitudes in the next step, and the finally choosen one in $V_{g}$\\
\hline
$\xi$ &tolerance of the difference between the end of the predicted trajectory and next waypoint &$t_{max}$&upper bound of the time which can be used to finish the predicted piece of trajectory \\
\hline
$a_{np},\ t_{np}$ & the acceleration and time variables for the optimization &$v_{n+1},\ p_{n+1}$& the predicted UAV velocity and position at the next step\\
\hline
$v_{min}$, $v_{max}$, $a_{max}$  & kinematic constraints for speed and acceleration & $p_{0},\ e_{0}$ & the position \& orientation message output by the buffer\\
\hline
$v_{p},\ v_{e}$ &the UAV translational and angular velocity for alignment&$t_{c},\ t_{p}$ &timestamps of $Pcl_2$ and the pose message of UAV \\
\hline
$a_p$, $a_e$ &the UAV translational and angular acceleration for alignment &$\Delta t_{n}$& average time cost of the motion planner loops\\
\hline
$t_{pm}$ & timestamp gap between UAV pose and current time & $ t_{dp}$ & timestamp gap between UAV pose and $Pcl_2$ \\
\hline
$t_{pl}$ & time cost of the former step of the motion planner&  $t_{ct}$ &fixed time cost for the flight controller\\
\hline
\end{tabular}
\end{center}
\vspace{-0.7cm}
\end{table*}
Our proposed framework is composed of three submodules that run parallelly and asynchronously: the point cloud filter (section \uppercase\expandafter{\romannumeral4.}A), the moving obstacle position and velocity estimator (section \uppercase\expandafter{\romannumeral3.}A \& B), and the motion planner (section \uppercase\expandafter{\romannumeral3.}C \& D). Their execution frequency is different, and the alignment for the message conveying between them is discussed in section \uppercase\expandafter{\romannumeral4}. Table \ref{table_1} explains the definitions of part of the symbols in this paper, other symbols are explained in (or near) the figures and algorithms for reading convenience.

\subsection{Classification for Static and Dynamic Obstacle}
First, the raw point cloud $Pcl_1$ is filtered to obtain $Pcl_2$, $p_{ali}$ and $e_{ali}$ are aligned to the timestamp of $Pcl_2$. $Pcl_2$ is the filtered point cloud in earth coordinate $E\!-\!X\!Y\!Z$, the details about the filter are in section \uppercase\expandafter{\romannumeral4}. The next step is to sort the obstacles by comparing two $Pcl_2$ frames at different moments and estimate the displacement of the obstacles. The detailed procedure is introduced in \textbf{Algorithm \ref{alg2}}. Its time complexity is $O(NlogN)$ to $O(N^2)$ and space complexity is $O(N)$. $N$ indicates the size of $Pcl_2$. First, for all the useful data, two data frames at the the former timestamp $t_1$ and latest timestamp $t_2$ is prepared (\textbf{Algorithm \ref{alg2}}, Line 2-3). To deal with the movement of the camera between the two frames, the core idea is filtering $Pcl_{t_2}$, only keeping the points within the FOV of the camera at $t_1$ (Line 4). So that the two point cloud frames can be regarded as from one static camera. It is important for the obstacle matching. The newly appeared obstacles in the latest frame are removed, so only the obstacles appear in both of the two point cloud frames are further analyzed. The minimal time interval $t_{2}-t_{1}$ is set to be able to make the displacement of the dynamic obstacles obvious enough to be observed, while maintaining a reasonable output frequency of the estimation results.


Next, the frame from $Pcl_2$ is separated into several clusters with each cluster representing one obstacle (Line 5). Density-based spatial clustering of applications with noise (DBSCAN) \cite{c27} can cluster the points into different obstacles by the density of the points and no pre-assigned cluster amount is required. The position of each obstacle is estimated by the spherical hypothesis, which is introduced in the next section. 

Then, the same obstacle in the two frames should be matched. This is matched by comparing the feature vector $fte()$ of two clusters, which is defined in (1). The feature vector contains six features that can be extracted from the point cloud, which are easy to calculate and proved to be very effective in tests. The idea is: if there is not a significant difference in the position, shape, and color of the two point clusters extracted from two point cloud frames respectively, then they are commonly believed the same object. 
$$ fte(ot_1) = [c_{1} , len(ot_{1}) , D(ot_{1}), V(ot_{1}), C_{M}(ot_{1}),C_{V}(ot_{1})] \eqno{(1)}$$

Finally, the moving obstacles are detected by comparing the obstacle center displacement with the threshold $d_s$ (\textbf{Algorithm \ref{alg2}}, line 8-9). $d_{Ecd}$ is the threshold for the feature vector difference, $d_{fte}<d_{Ecd}$ stands for that the same obstacle is matched. The points of the static obstacle are stored in a memory list $Pcl_s$ with a fixed size, to remember the static obstacles that just move out the FOV. $C_d$ is cleared at the beginning of each loop, while $Pcl_s$ not.

\vspace{-0.3cm}
\begin{algorithm}[!h]
\caption{Sorting the static and dynamic obstacles}
\label{alg2}
\begin{algorithmic}[1]
\WHILE{true:}
\STATE Record $Pcl_2$, $p_{ali}$, $e_{ali}$ with their timestamps into separate lists
\STATE Extract $Pcl_{t_1}$,$Pcl_{t_2}$,$p_{t_1}$,$p_{t_2}$,$e_{t_1}$,$e_{t_2}$ at timestamp $t_{1},t_{2}$ $(t_{2}-t_{1}>d_{t})$ from the lists built in line 2
\STATE Remove points in $Pcl_{t_2}$ which is out of the FOV of camera at the pose at $t_1$
\STATE Cluster $Pcl_{t_1},Pcl_{t_2}$ with DBSCAN to obtain the clusters $OT_{1}(ot_{11},ot_{12},...)$, $OT_{2}(ot_{21},ot_{22},...) $
\STATE Calculate the corresponding obstacle position (section \uppercase\expandafter{\romannumeral4.}B), get $C_{1}(c_{11},c_{12},...)$ and $C_{2}(c_{21},c_{22},...)$
\FOR {$ot_{2k}$ in $OT_2$ ($k$ is the iteration number):}
\STATE Calculate the Euclidean distance $d_{fte}=\|fte(ot_1)-fte(ot_{2k})\|_{2}$ with all $OT_{1}$ and $ot_{2k}$, $ot_{1j}$ ($j$ is the index) in $OT_{1}$ is corresponding to the minimal distance, $d_{kj} = \overline {c_{2k}c_{1j}}$.
\IF{$d_{kj}<d_{s}$ and $d_{fte}<d_{Ecd}$:}  
\STATE Add $ot_{2k}$ to list $Pcl_{s}$
\ELSE
\STATE Add $c_{2k}$ to list $C_{d}$
\ENDIF
\ENDFOR
\ENDWHILE
\end{algorithmic}
\end{algorithm} 
\vspace{-0.3cm}
%


\subsection{Obstacle Position and Velocity Estimation}

The velocity $v_{2k}$ of the obstacle $ot_{2k}$ can be obtained by $ v_{2k} = \overrightarrow {c_{2k}c_{1j}}/{(t_{2}-t_{1})}$. Besides, a Kalman filter in (2) is adopted to further reduce the position and velocity estimation error and make the estimating output denser. The superscript $^-$ indicates the prior state matrix $\hat x_t$ and the error covariance matrix $P_{t}$ before updated by the Kalman gain matrix $K_t$. $F_{t}$ is the state transition matrix and $B_t$ is the control matrix. $x_t$ is composed of the obstacle position and velocity before the filtering, and $\hat \ $ marks the filtered results for $x_t$. $\hat{x}_{t}$ equals to the predicted state $\hat{x}_{t}^{-}$ if no dynamic obstacle is 
caught by \textbf{Algorithm \ref{alg2}}. The subscripts $t$ and ${t\!-\!1}$ distinguish the current and former step of the Kalman filter. $\Delta t$ is the time interval between each run of the Kalman filter. On the current stage, we assume the moving obstacle performs uniform motion between $t_1$ and $t_2$, $a_{t-1}=0$.




\vspace{-0.4cm}
$$\begin{array}{c}
\hat{x}_{t}^{-}=F_{t} \hat{x}_{t-1}+B_{t} a_{t-1},\quad P_{t}^{-}=F_{t} P_{t-1} F_{t}^{T}+Q\\
K_{t}=P_{t}^{-} H^{T}\left(H P_{t}^{-} H^{T}+R\right)^{-1},\quad P_{t}=\left(I-K_{t} H\right) P_{t}^{-}\\
\hat{x}_{t}=\left\{\begin{array}{l} \hat{x}_{t}^{-}+K_{t}\left(x_{t}-H \hat{x}_{t}^{-}\right) \ \text{($C_d$  is not empty)} \\
\hat{x}_{t}^{-}\  \text{($C_d$ is empty)}
\end{array}\right.\\
x_{t}= \left[\begin{array}{l}
c_{2k} \\
v_{2k}
\end{array}\right],F_{t}=\left[\begin{array}{cc}
1 & \Delta t \\
0 & 1
\end{array}\right], B_{t}=\left[\begin{array}{c}
\frac{\Delta t^{2}}{2} \\
\Delta t
\end{array}\right]
\end{array}\eqno{(2)} $$

\subsection{Collision check and velocity planning with the relative velocity method}

\subsubsection{Review of the HAS method}
\begin{figure}[!h]
      \centering
      \includegraphics[width=0.40\textwidth]{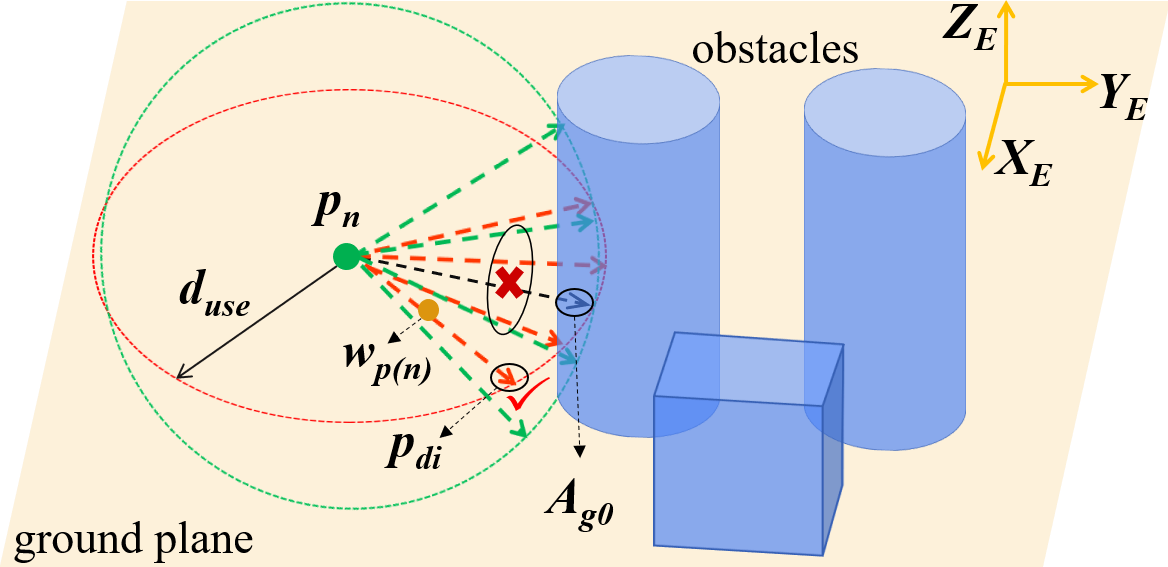}
      \caption{The illustration of HAS process. The red dash circle and red arrows are for the search on the plane parallel to ground, while the green ones are for the vertical plane.}
      \label{fig-has}
      \vspace{-0.0cm}
   \end{figure}

To help understand \textbf{Algorithm \ref{alg5}}, our previously proposed HAS method \cite{c26} is briefly revisited as follows. In HAS, a waypoint $w_{p(n)}$ close to the drone is first searched as the trajectory endpoint constrain for the motion planning. Moreover, the real trajectory between $p_n$ and $w_{p(n)}$ is proved collision-free. As shown in Fig. \ref{fig-has}, a bunch of line segments spread out from the initial search direction $A_{g0}$, and these line segments have a common start point $p_n$ and the same length $d_{use}$. $d_{use}$ is the point cloud distance threshold, the points whose distance from $p_{n}$ further than $d_{use}$ are not considered in the collision check. The two line segments of symmetry about $A_{g0}$ on the plane parallel to the ground plane are first checked if they collide with obstacles. If they collide, the two lines in the vertical plane will be checked. These four lines have the same angle difference with $A_{g0}$. If the minimal distance between the line and obstacles is smaller than $r_{safe}$, it is treat as collision. Fig. \ref{fig-has} shows when the first round of search has failed, another round with greater angle difference is conducted until a collision-free direction $\overrightarrow{p_{n}p_{di}}$ is found. $w_{p(n)}$ is on $\overrightarrow{p_{n}p_{di}}$ and $\|p_{n}w_{p(n)}\|_2$ should satisfy the safety analysis in \cite{c26}.
\vspace{-0.2cm}

\subsubsection{Collision check and velocity planning}

\begin{figure}[h]
\centering
\subfigure[]{
\includegraphics[width=0.29\textwidth]{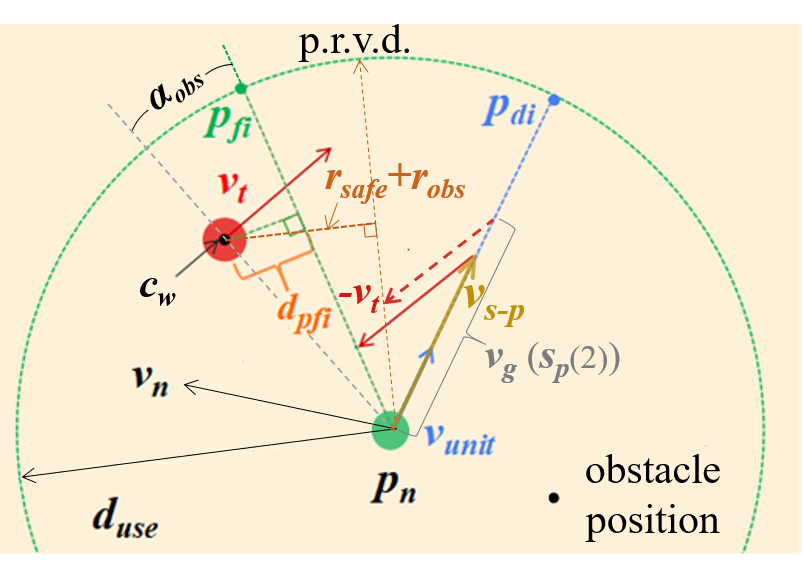}}
\hfill
\centering
\subfigure[]{
\includegraphics[width=0.17\textwidth]{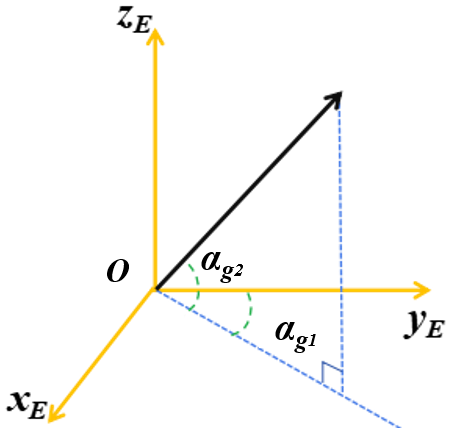}}
\caption{The graph illustration for the relative velocity method. To be clear, only one moving obstacle is shown in the figure (a). $v_t$ is the obstacle velocity. At first $d_{pfi} < r_{obs} + r_{safe}$. After \textbf{Algorithm \ref{alg5}}, the planned velocity magnitude $v_g$ is determined. ``p.r.v.d'' stands for the finally planned relative velocity direction. The distance from $p_{n}$ to p.r.v.d equals $ r_{obs} + r_{safe}$. It is also safe if the drone speed is greater than $v_g$. Figure(b) depicts the two return components $\alpha_{g1}$ and $\alpha_{g2}$ of function ag(), i.e., $ag()=(\alpha_{g1},\alpha_{g2})$.}
\label{fig6}
\vspace{-0.1cm}
\end{figure}
\setlength{\textfloatsep}{0.1cm}
\begin{algorithm}[!h]
\caption{collision check and velocity planning} 
\label{alg5}
\begin{algorithmic}[1]

\STATE Run \textbf{Algorithm 4} (HAS) in \cite{c26} with $Pcl_s$, find the collision-free $\overrightarrow{p_{n}p_{di}}$ for static obstacles.
\FOR{$c_{w}$ in $\hat C_d$ ($w$ is the iteration number):}
\STATE $d_{pfi}=\dfrac{\|\overrightarrow{p_{n} c_{w}} \times \overrightarrow{c_{w} p_{fi}}\|_{2}}{\left\|\overrightarrow{p_{n} p_{fi}}\right\|_{2}}$
\IF {$d_{pfi} < r_{obs} + r_{safe}$ $\wedge$ $\angle c_{w}p_{n}p_{fi} < 90^{\circ}$ }
\STATE $\alpha_{obs} = arcsin\left(\dfrac{r_{obs}+r_{safe}}{\left\|\overrightarrow {p_{n}c_{w}}\right\|_{2}}\right)$
\STATE $\Gamma(m+1)=(rd((4-m)/4),rd(m/4))\alpha_{obs}+ag(\overrightarrow {p_{n}c_{w}})$, $m=\{0,1,2,3\}$, $\alpha_{0}=ag(\overrightarrow {p_{n}p_{fi}})$
\STATE $\gamma = \Gamma(argmin(\|\Gamma(1)- \alpha_{0}\|_{2},...,\|\Gamma(4)-\alpha_{0}\|_{2}))$
\STATE $s_{p} = v_{t} \cdot A_{t}$
\IF {$ \| v_{unit}s_{p}(2)- v_{n}\|_{2}  < \Delta t_{n} a_{max}$}
\STATE Add $s_{p}(2)$ to list $V_g$
\ELSE
\STATE Break the circle
\ENDIF
\ENDIF
\ENDFOR
\IF {$V_g$ is not empty}
\STATE Convert $V_{g}$ to vector, $v_{g} = V_{g}(argmin(abs(V_{g}- \|v_{n}\|_{2})))$
\ENDIF
\end{algorithmic}
\end{algorithm} 
\vspace{-0.0cm}
$$v_{unit}=\dfrac{\overrightarrow{p_{n} p_{d i}}}{\left\|\overrightarrow{p_{n} p_{di}}\right\|_{2}} ,
A_{t}= \left[\begin{array}{cc}
cos(\gamma(1)) & v_{unit}(1) \\
sin(\gamma(1)) & v_{unit}(2) \\
tan(\gamma(2)) & v_{unit}(3)
\end{array}\right]\eqno{(3)}$$




\textbf{Algorithm \ref{alg5}} reveals the process of searching for the next desired velocity for avoiding multiple moving obstacles. Its time and space complexity are both $O(N)$. (3) gives the definition for $v_{unit}$ and $A_{t}$. $v_{s-p}$ is the original planned velocity if only HAS is used, $v_{s-p}$
is parallel to $\overrightarrow{p_{n}p_{di}}$ and $\|v_{s-p}\|_{2}=\|v_{n}\|_{2}$. $\overrightarrow {p_{n}p_{fi}}$ is parallel to the relative velocity $(v_{s-p}-v_{t})$. Fig. \ref{fig6}(a) shows the situation when the planned velocity $v_g$ is found by \textbf{Algorithm \ref{alg5}}. 

At the beginning, the original collision check algorithm of HAS method \cite{c26} is used, and \textbf{Algorithm \ref{alg5}} mainly introduces the specific process of collision check based on the relative velocity method. The direction angle function $ag()$ is explained in Fig. \ref{fig6}(b), it returns a 2d vector. If $v_{s-p}$ is not safe ($d_{pfi} < r_{obs} + r_{safe}$), we calculate the direction angle of $\overrightarrow{p_{n}c_{w}}$, and add $\alpha_{obs}$ from four orthogonal directions on it to obtain 4 feasible directions of the relative velocity (FDRV). $\alpha_{obs}$ is the minimum 
angle increment to ensure the FDRV keeps a safe distance from the obstacle $c_{w}$. $\angle c_{w}p_{n}p_{fi} < 90^{\circ}$ is to ignore the obstacles who are ``behind'' the drone because they have no collision risk. $\Gamma$ is the list for storing the direction angle. Line 7 finds the FDRV with the minimal acceleration cost for the current obstacle $c_{w}$, denoted as $\gamma$. Because only 2 vertical and 2 horizontal directions are considered, the acceleration to reach the planned velocity is only near optimal. Line 9 checks if the desired speed can be reached within the acceleration constrain. Finally, Line 17 finds $v_g$ with the minimal acceleration from all the feasible velocity magnitudes. If $v_g$ can not be found after \textbf{Algorithm \ref{alg5}}, the current search direction $\overrightarrow{p_{n}p_{di}}$ fails, and \textbf{Algorithm \ref{alg5}} starts with the next search direction in HAS. If there is no moving obstacle, $\|v_g\|_2$ defaults to equal $v_{max}$.



In conclusion, \textbf{Algorithm \ref{alg5}} calculates the end of the speed range for the drone at the next step for the motion optimization. It is a more precise and energy-saving way to control the drone rather than suddenly brake down or accelerate. The safe speed range is introduced in section \uppercase\expandafter{\romannumeral3.}D. In most cases of simulation tests, \textbf{Algorithm \ref{alg5}} can be done within 9 ms on average.
 

\vspace{-0.1cm}
\subsection{Motion Planning}
After obtaining the waypoint $w_{p(n)}$ and desired velocity magnitude boundary $v_g$, the motion primitives are calculated and sent to the flight controller. The optimization problem to obtain motion primitives is defined in (4). $v_{n+1}$ and $p_{n+1}$ are calculated by the kinematic formula. $\eta_{1},\ \eta_{2}$ are coefficients, the default values are shown in Table \ref{table_2}. 

The constraints and the objective function are different by whether the dynamic obstacle exists. If there exists a moving obstacle, $\xi = \infty $ to convert the trajectory end point constrain to the component of object function. Otherwise $\eta_{2} =0$ to strictly constrain the trajectory end point. If $\|v_{g}\|_{2} < \|v_{n}\|_{2}$ then $v_{max} = \|v_{g}\|_{2}$, otherwise $v_{min} = \|v_{g}\|_{2}$. This is the safe speed range, it decides the drone should whether decelerate or accelerate according to the planned speed. In any above conditional equation, the parameter not mentioned equals to the default value given in Table \ref{table_2}. The constraint $a_{max}$ ensures the aircraft can fly within its own kinematic limit. The object function is designed to solve the minimum acceleration and the shortest time cost to reach the next waypoint and desired velocity. By adjusting $\eta_{1}$, the flight pose stationarity and the speed can be balanced. The drone will fly more aggresively if we increase $\eta_{1}$.
 
 
 \vspace{-0.4cm}
$$\begin{aligned}
& \min _{a_{np}, t_{np}}\left\|a_{np}\right\|_{2} +\eta_{1} t_{np}+\eta_{2} \| \overrightarrow{p_{n+1}w_{p(n)}} \|_{2}\\
\text{s.t.}\ &0< \mathrm{t}_{np} \leq t_{\max },\  v_{min} \leq \| v_{n+1}\|_{2}\leq  v_{max}\\
&\left\|a_{np}\right\|_{2} \leq a_{\max },\ \| \overrightarrow{p_{n+1}w_{p(n)}} \|_{2} \leq  \xi \\
&v_{n+1}=v_{n}+a_{np} t_{np},\ p_{n+1}=p_{n}+v_{n} t_{np}+\frac{1}{2} a_{np} t_{np}^{2}
\end{aligned} \eqno{(4)}$$

\vspace{-0.1cm}
\section{IMPLEMENTATION OF THE METHODS}
This section introduces the technique details for implementing and improving our proposed method.
\vspace{-0.1cm}
\subsection{Raw point cloud filtering}

\begin{figure}[thpb]
      \centering
      \includegraphics[width=0.43\textwidth]{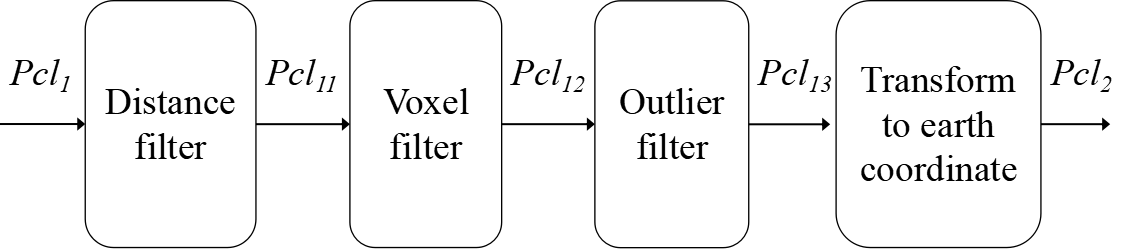}
      \caption{The filtering process from $Pcl_1$ to $Pcl_2$}
      \label{fig3}
      \vspace{-0.2cm}
   \end{figure}

In this paper, filtering the raw point cloud is necessary, because the obstacle estimation is sensitive to the noise. The noise should be eliminated strictly, even losing a few true object points is acceptable. The filter has the same structure with our former work \cite{c26}, as shown in Fig. \ref{fig3}, but the parameters are different. $Pcl_{11}$, $Pcl_{12}$, $Pcl_{13}$ are the output of the three filters in Fig. \ref{fig3} repectively. Distance filter removes the points too far ($\geq 4\ m$) from the camera, voxel filter keeps only one point in one fixed-size ($0.2\ m$) voxel, outlier filter removes the point that does not have enough neighbors ($\leq$ 13) in a certain radius ($0.3\ m$). These metrics are tuned manually during extensive tests on the hardware platform introduced in section \uppercase\expandafter{\romannumeral5}, to balance the point cloud quality and the depth detect distance. They are proved satisfactory for obstacle position estimation. The point cloud filtering also reduces the message size by one to two orders of magnitude, so the computation efficiency is much improved, while the reliability of the collision check is not affected.

However, when the drone performs an aggressive maneuver, the motion blur of the camera may cause the wrong point cloud output, which may fail the collision-free waypoint generation. To solve this problem, we propose a practical and effective measurement: Only the filtered point cloud $Pcl_2$ is accepted when the angular velocity of the three Euler angles of the drone is within the limit $\omega_{max}$. 

In addition, before \textbf{Algorithm \ref{alg2}} is executed, the pose feedback message from the UAV controller should align with the timestamp on $Pcl_2$. It is to convert $Pcl_{13}$ in the body coordinate $B\!-\!xyz$ more precisely to $Pcl_2$ in the earth coordinate $E\!-\!X\!Y\!Z$. So a message buffer is set to output the aligned point cloud and UAV pose message within a short time interval (i.e. $\leq 0.03\ s$). After that the pose of UAV at the timestamp of $Pcl_2$ is further estimated in (5) by kinematics formulation. In (5) we assume that the 
translational and rotational acceleration are constant during time ${t}_{c}\!-\!{t}_{p}$, the accuracy is satisfactory when ${t}_{c}\!-\!{t}_{p}$ is small enough.
\vspace{-0.3cm}
$$\left[\begin{array}{c}{p}_{ali}\\ {e}_{ali}\end{array}\right]=\left[\begin{array}{c}{p}_{0}\\ {e}_{0}\end{array}\right]+\left[\begin{array}{c}{v}_{p}\\ {v}_{e}\end{array}\right]\left({t}_{c}-{t}_{p}\right)+\dfrac{1}{2}\left[\begin{array}{c}{a}_{p}\\ {a}_{e}\end{array}\right]\left({t}_{c}-{t}_{p}\right)^{2}\eqno{(5)}$$

\subsection{Spherical hypothesis}
   
\begin{figure}[h]
      \centering
	\subfigure[]{
      \includegraphics[width=0.29\textwidth]{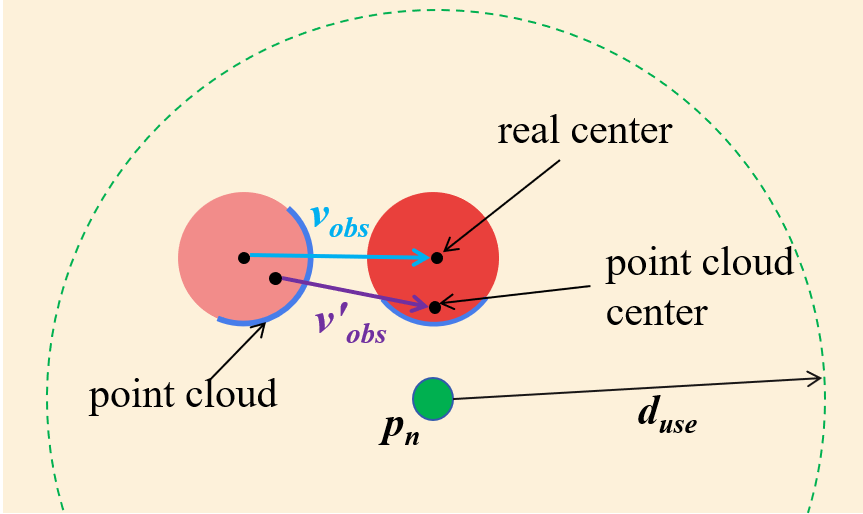}}
    \centering
	\subfigure[]{\includegraphics[width=0.18\textwidth]{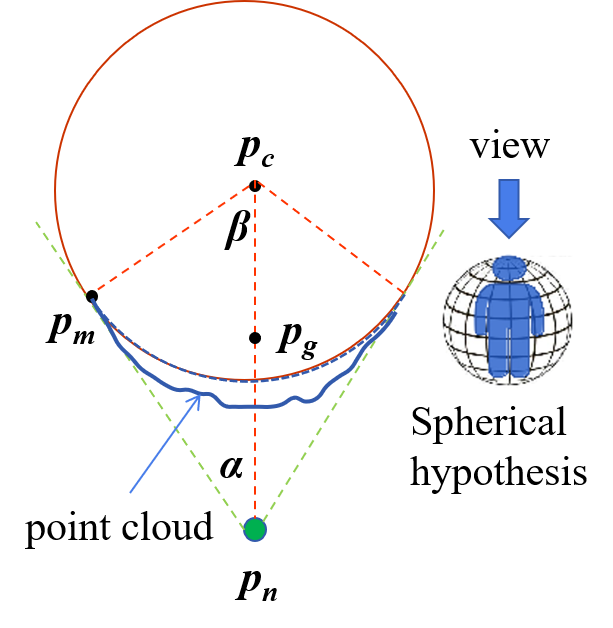}}
      \caption{The error in estimating the obstacle position and the spherical hypothesis. The real situation is 3D. In order to facilitate the presentation of the concept, it is put on a floor plan, and it is the same for Fig. \ref{fig6}.}
      \label{fig4}
      \vspace{-0.3cm}
   \end{figure}

    
The point cloud only describes a part of the obstacle which is facing the camera, as shown in Fig. \ref{fig4}(a). $v_{obs}$ is the real velocity of the obstacle, while $v'_{obs}$ is the estimated velocity with the point cloud center. The point cloud center position has obvious difference between the real obstacle center, so it can not be used as the obstacle position. Thus, the spherical hypothesis is proposed to reduce the error by supposing the obstacle as a sphere. 
In Fig. \ref{fig4}(b), $p_{c}$ is the center of the sphere, $p_{g}$ is the point cloud center and its coordinate is known. $p_{m}$ is the farthest point in the point cloud from $p_{g}$. $\overrightarrow{p_{n}p_{m}}$ is the tangent of the circle. $r_{obs}=\overline {p_{c}p_{m}}$ is the radius of the sphere. The coordinate of $p_{c}$ can be solved by (6), it is the obstacle position used in \textbf{Algorithm \ref{alg5}}.

  \vspace{-0.2cm}

 \vspace{-0.4cm}
$$ \begin{aligned} 
\alpha = arccos(\dfrac{ \overline {p_{m}p_{n}}^2+ \overline {p_{g}p_{n}}^2- \overline {p_{m}p_{g}}^2}{2 \overline {p_{m}p_{n}} \cdot \overline {p_{g}p_{n}}})\\
\beta = \dfrac{\pi}{2}- \alpha \text{ , } \overline {p_{c}p_{m}}  =  \overline {p_{m}p_{n}} tan(\alpha)\\
 \overline{p_{c}p_{g}} =  \overline {p_{c}p_{m}} \dfrac{sin(\beta)}{\beta} \text{ , } p_{c} =p_{n} + \overrightarrow {p_{g}p_{n}}\dfrac{ \overline {p_{c}p_{n}} }{ \overline {p_{g}p_{n}}}
\end{aligned}\eqno{(6)}$$

Although the spherical hypothesis also causes errors because an obstacle in real-life is rare to be a perfect sphere, it is reliable in safety. The estimation errors of the obstacle position and radius will not harm the safety with the relative velocity method. Intuitively, because the backside of an obstacle is not considered in the collision check, only the part can be seen by the camera matters the collision check \cite{c26}. In Fig. \ref{fig4}, the projected area of the proposed sphere on camera is not smaller than the point cloud. Thus, it is safe to replace the point cloud with this sphere.

\vspace{-0.2cm}
\subsection{Position estimation for deformable obstacles}
Here we introduce another technique for reducing the position estimation error of obstacle if the obstacle deforms during its traveling, such as a walking human. The waving limbs of a walking human often increase the human position estimation error due to the point cloud deformation. But when two neighbor frames of $Pcl_2$ are overlaid (i.e. $Pcl_{t_2}=Pcl_{t_2}\cup Pcl_{t_{2}-b}$, the subscript $_{-b}$ refers to the former neighboring message), the point cloud of the human trunk is denser than the other parts which rotate over the trunk.  Then increasing the point density threshold of DBSCAN can remove the points corresponding to the limbs. The mean of $(p_{t_2},\ p_{t_{2}-b})$ is used to replace $p_{t_2}$. For example $p_{t_2}=(p_{t_2}+p_{t_{2}-b})/2$, and it is the same for $p_{t_1},e_{t_1}$ and $e_{t_2}$ (\textbf{Algorithm \ref{alg2}}, line 3).
\subsection{Position prediction in the planner}
The motion of the drone is more aggressive for avoiding moving obstacles than flying in a static environment. To address the displacement of the drone during the time costed by the trajectory planner and flight controller, time compensation is adopted before implementing \textbf{Algorithm \ref{alg5}}. The current position of drone $p_{n}$ is updated by the prediction, as shown in (7). In addition, due to the time cost of obstacle identification, the timestamp on the information of dynamic obstacles is always delayed than that of the pose and velocity message of the drone. Based on the uniform 
acceleration assumption, the obstacle position coordinate $\hat c_{2k}$ at the current time is predicted and updated with the time gap $t_{dp}$ as (8).

\vspace{-0.4cm}
$$ p_{n} = p_{n} +( t_{pl} + t_{ct} + t_{pm})v_{n} + \frac{1}{2}( t_{pl} + t_{ct} + t_{pm})^2 a_{n}
\eqno{(7)}$$
$$ \hat c_{2k} = {\hat c}_{2k} + ( t_{pl} + t_{ct} + t_{pm} + t_{dp} )\hat v_{2k}
\eqno{(8)}$$

\section{EXPERIMENTAL RESULTS}
\subsection{Experimental Configuration}

The detection and avoidance of obstacles are tested and verified in the Robot Operation System (ROS)/Gazebo simulation environment first and then in the hardware experiment. The drone model used in the simulation is 3DR IRIS, and the underlying flight controller is the PX4 1.10.1 firmware version. The depth camera model is Intel Realsense D435i with resolution 424*240 (30 fps). For hardware experiments, we use a self-assembled quadrotor with a QAV 250 frame and an UP board with an Intel Atom x5 Z8350 1.44 GHz processor, other configuration keeps unchanged. Table \ref{table_2} shows the parameter settings for the tests. The supplementary video for the tests has been uploaded online\footnote{https://youtu.be/TgHY7i5K6v8}.
\vspace{-0.2cm}
\begin{table}[h]
\caption{parameters for the simulation}
\vspace{-0.4cm}
\label{table_2}
\begin{center}
\begin{tabular}{|c|c|c|c|c|c|}
\hline
Parameter& Value &Parameter & Value & Parameter & Value\\
\hline
$d_{use}$ &3 m & $t_{ct}$ &0.01 s & $a_{max}$ & 4 m/s$^2$\\
\hline
$ t_{max}$ & 1.2 s &$v_{max}$ &3 m/s & $r_{safe}$ &0.5 m\\
\hline
$ \omega_{max}$ & 1.5 rad/s & $\overline{p_{n}w_{p(n)}}$ & 0.3m &$ t_{2}-t_{1} $ & 0.2 s \\
\hline
$\eta_1$ & 6 &$\eta_2$ & 20 &$\xi$ & 0.02\\
\hline
\end{tabular}
\end{center}
\vspace{-0.6cm}
\end{table}

\subsection{Simulation Test}

First, the accuracy and stability of the estimation method for the obstacle position and velocity are verified. 

\begin{figure}[h]
\subfigure[]{
\includegraphics[width=0.2\textwidth , height = 3cm]{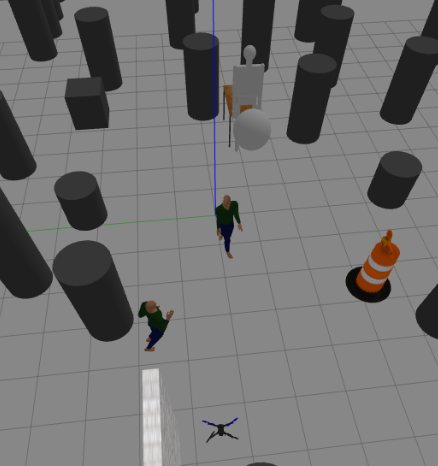}}
\hfill
\centering
\subfigure[]{
\includegraphics[width=0.2\textwidth , height = 3cm]{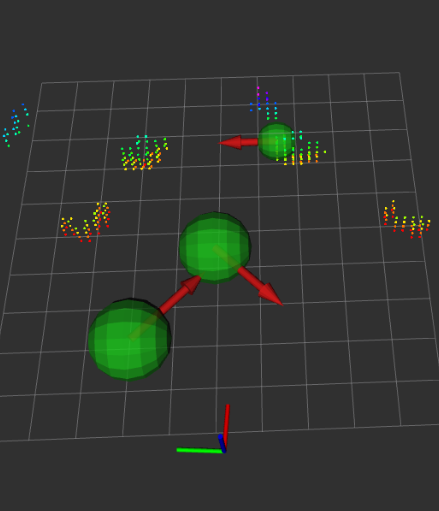}}
\caption{The simulation environment for the moving obstacles' position and velocity estimation test. The colorful points stand for static obstacles.}
\label{fig9}
\vspace{-0.3cm}
\end{figure}

\begin{figure}[h]
\centering
\subfigure[]{
\includegraphics[width=0.23\textwidth, height=3cm]{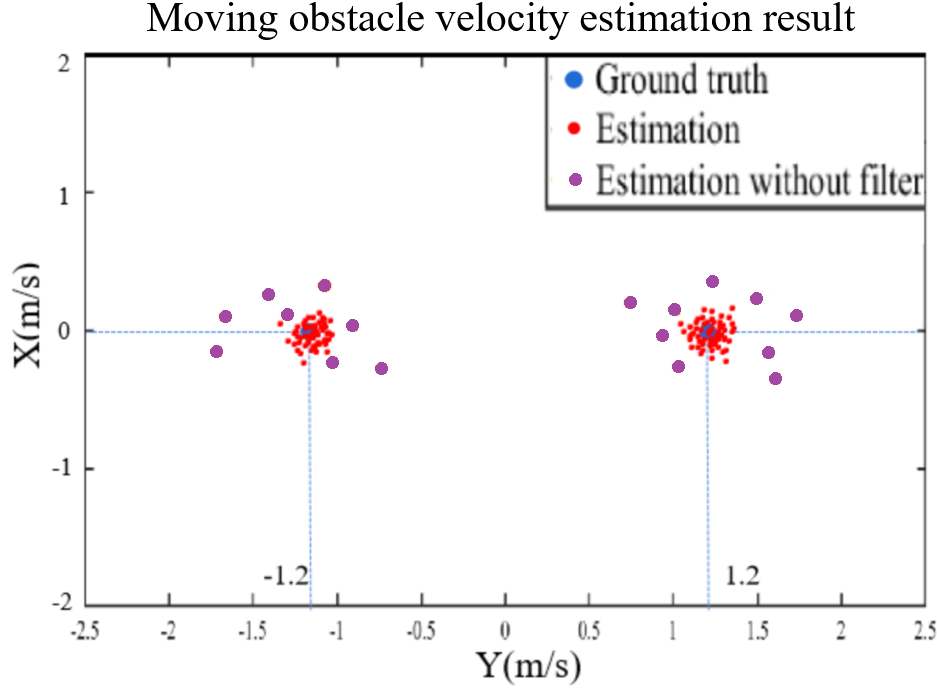}}
\hfill
\centering
\subfigure[]{
\includegraphics[width=0.23\textwidth, height=3cm]{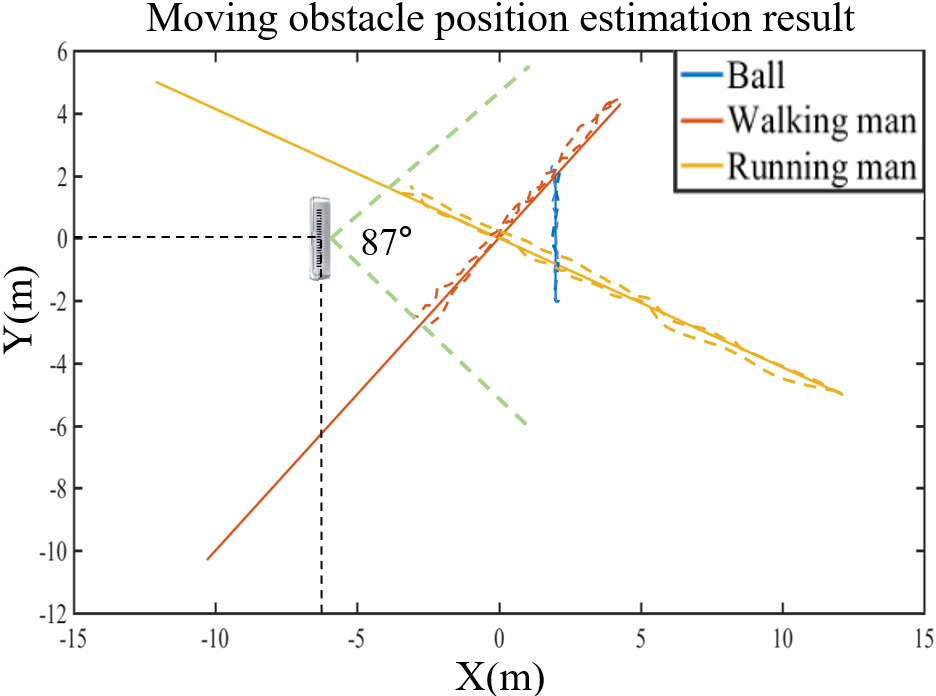}}
\caption{The estimation results of the moving obstacles' velocity and position}
\label{fig10}
\vspace{-0.3cm}
\end{figure}

In the simulation world as depicted in Fig. \ref{fig9}(a), there are one moving ball, two moving human models and some static objects. The camera is fixed on the head of the drone, facing forward straightly. The drone is hovering around the point $(-6,0,1.5)$ in the range of 1 $m$ to keep the camera moving. Fig. \ref{fig9}(b) depicts the visualized estimation results in Rviz. The green spheres indicate moving obstacles, the red arrows represent the obstacle velocity. The estimation numeral results are shown in Fig. \ref{fig10}. We compare them with the ground truth and the estimation results without the Kalman filter. Fig. \ref{fig10}(a) exhibits that the Kalman filter significantly improves the accuracy. The average velocity estimation error is reduced from $\pm 0.49\ m/s$ to $\pm 0.12\ m/s$, and the original data develops from sparse into dense. In Fig. \ref{fig10}(b), the dash and solid lines represent the estimated position and the ground truth of it. The average position estimation error for the moving person (0.31 m) is larger than the moving ball (0.12 m) due to body posture changes.
The estimation test results demonstrate that our estimation algorithm is practical for dynamic obstacle avoidance, the velocity estimation relative error is only about $10\%$. 

\begin{figure}[thpb]
\centering
\subfigure[]{
\includegraphics[width=0.22\textwidth,height = 3cm]{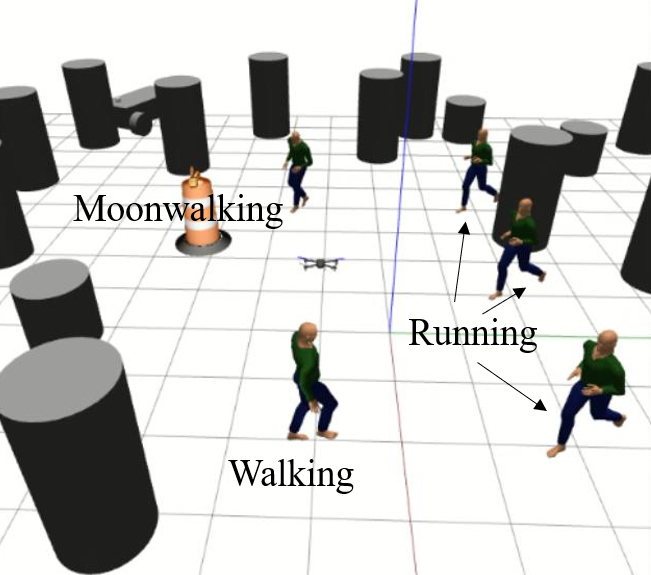}}
\hfill
\centering
\subfigure[]{
\includegraphics[width=0.24\textwidth,height = 3cm]{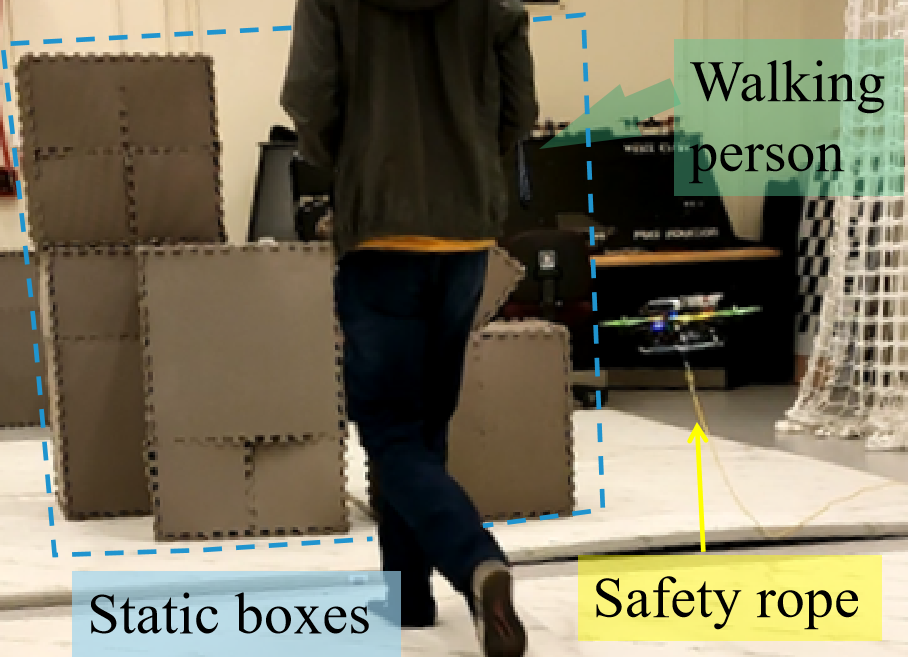}}
\hfill
\centering
\subfigure[]{
\includegraphics[width=0.49\textwidth]{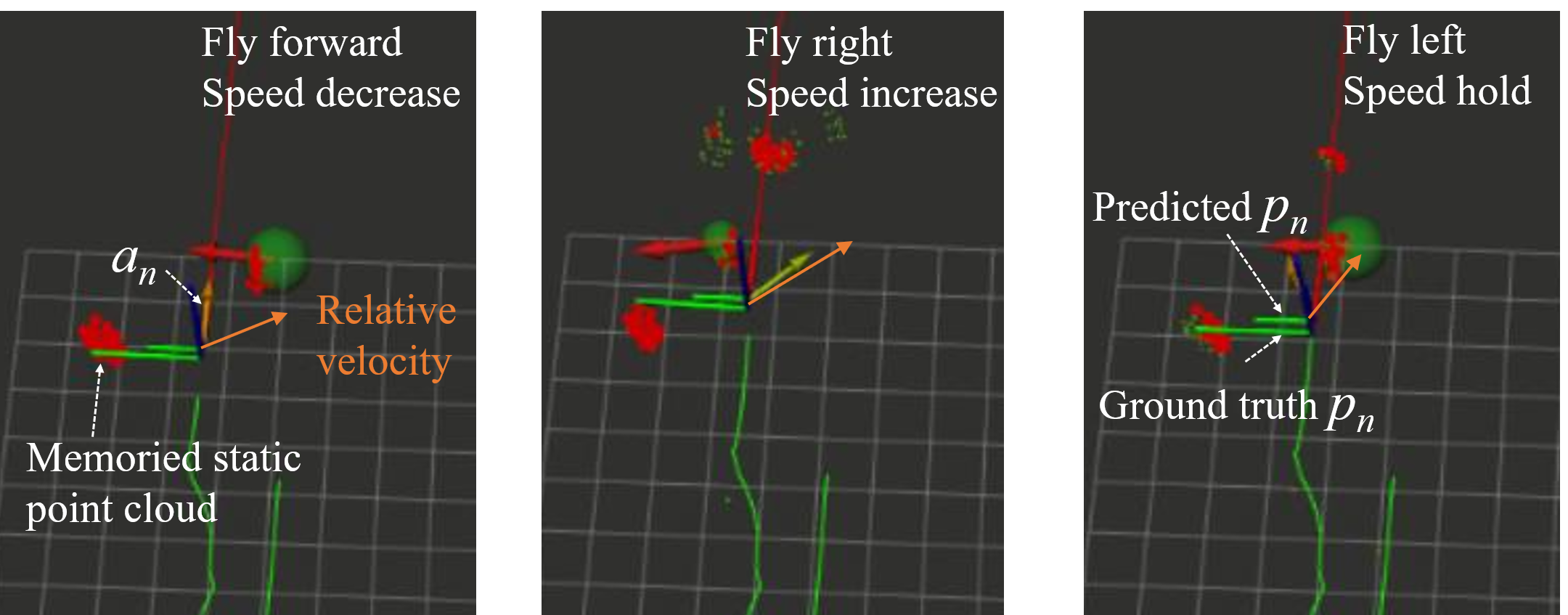}}
\centering
\subfigure[]{
\includegraphics[width=0.46\textwidth,height = 4.7cm]{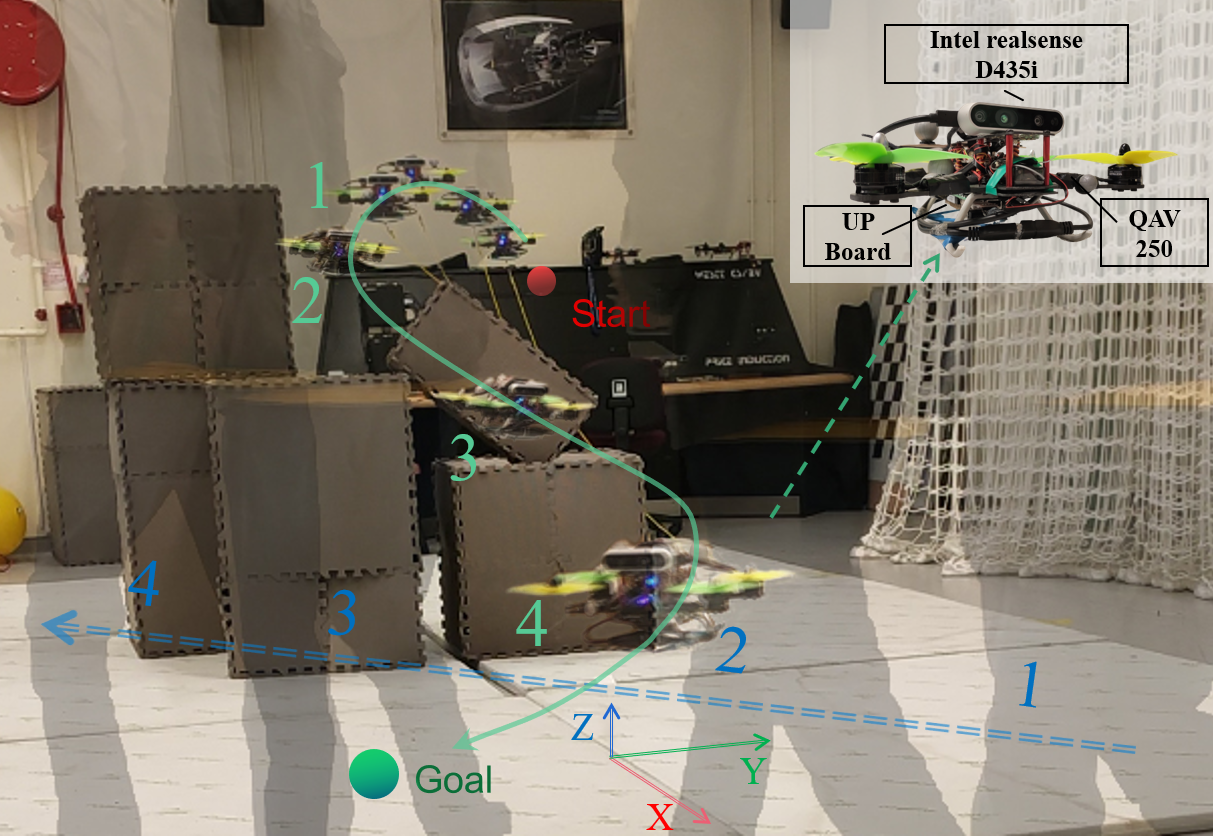}}
\vspace{-0.1cm}
\caption{(a): The dynamic simulation test environment. (b): The dynamic hardware test environment. (c): the data in RVIZ corresponding to Fig. \ref{fig1}. (d): The composed picture of the hardware flight test. Numbers in (d) mark the same frame for the drone and person.}
\label{fig11}
\vspace{-0.1cm}
\end{figure}

Fig. \ref{fig11}(a) is the overview of the simulation world. In Fig. \ref{fig1}, the better safety of our proposed method is illustrated: To avoid the moving man which is at a similar speed with the drone, the aircraft choose to fly in the ``opposite'' direction with the man so the threat is removed easily. If only the static HAS method is utilized in the same situation, the drone decelerates and fly alongside the man(red line), which is very inefficient. Fig. \ref{fig11}(c) is the visualized data in RVIZ to show different motion choices in an obstacle avoidance procedure with and without the relative velocity method (the left two \& the right figure). The data is processed after the flight simulation to calculate the relative velocity and draw it on the picture accordingly. The red straight line connects the goal 
and $p_{n}$, and the green line is the real path updated at 1 Hz. For the two left figures in Fig. \ref{fig11}(c), the drone detects a left-moving obstacle at its forward right side first. The drone chooses to decelerate to avoid it, but finally the drone finds the obstacle moves slower than what it was observed before. So the best choice is to move right, \textit{i.e.}, in the opposite moving direction of the obstacle. For the right figure, the drone fly left to choose the shortest path to pass, considering the person as static. But it probably leads to a collision, because the drone and person go in the same direction. All in all, with the relative velocity method, the orientation of relative velocity always keeps a safe distance from the moving obstacles. 

\vspace{-0.3cm}
\vspace{-0.4cm}


\vspace{-0.0cm}
  
\subsection{Hardware Test}
\begin{figure}[thpb]
      \centering
      \subfigure[]{
      \includegraphics[width=0.48\textwidth,height =2.7cm]{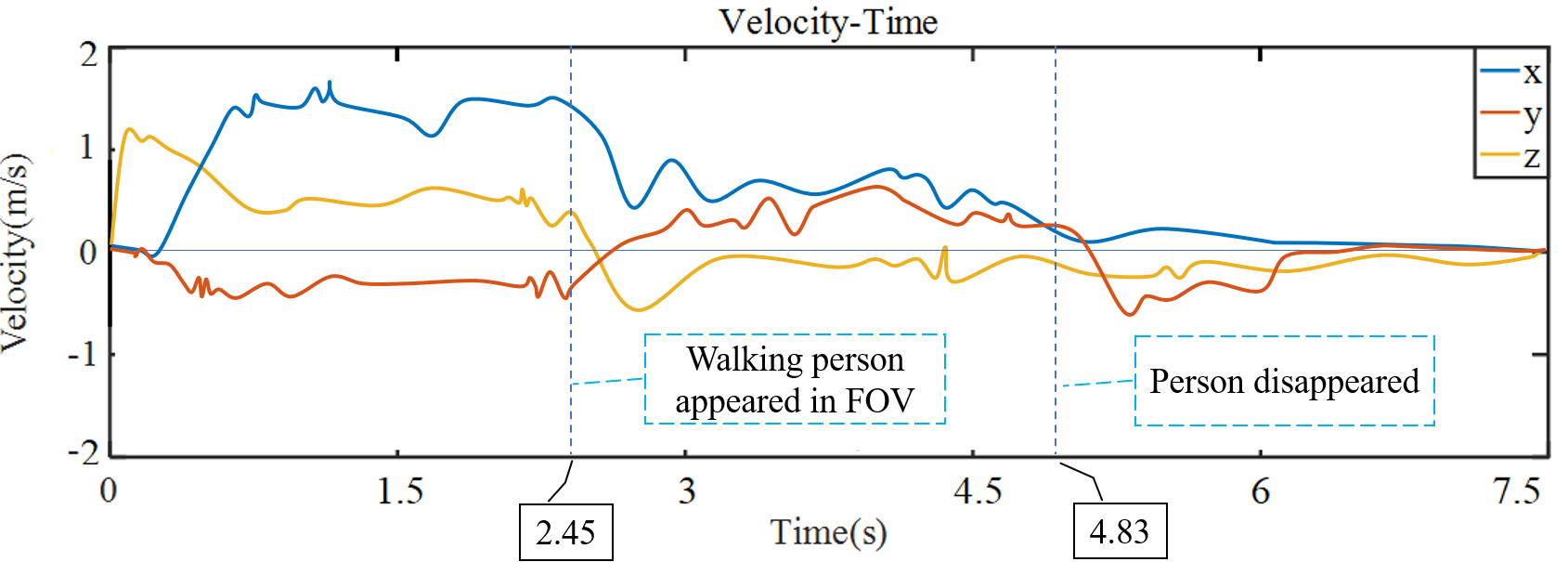}}
    \centering
      \subfigure[]{
           \includegraphics[width=0.49\textwidth,height =2.5cm]{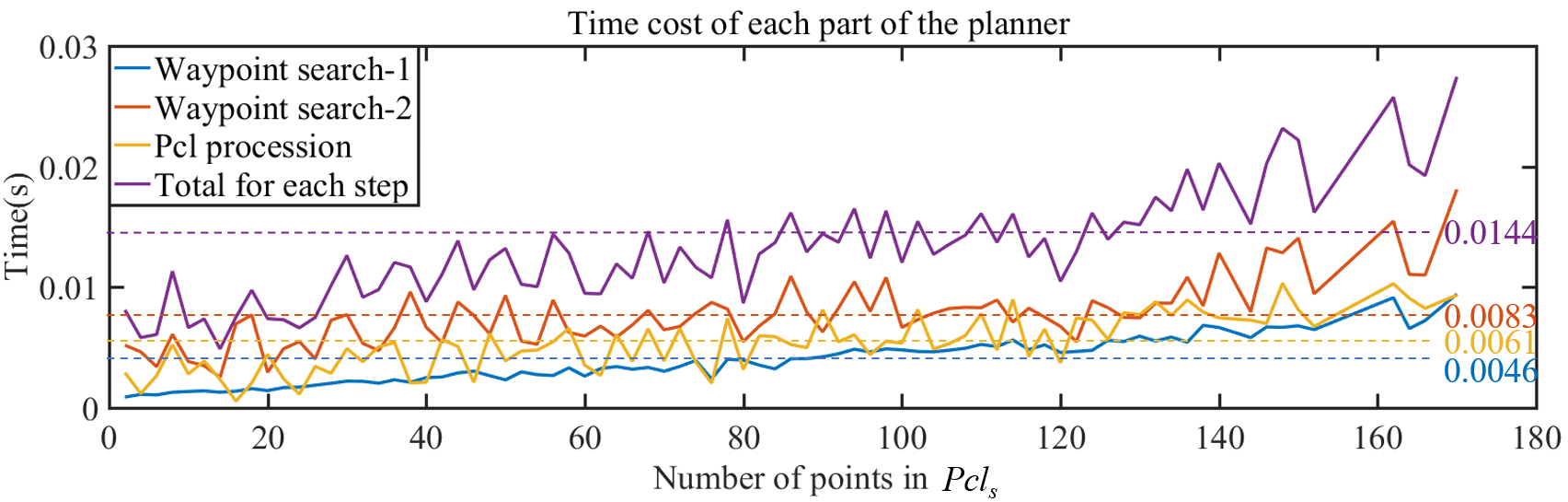}}
     \caption{(a): The velocity of the drone in earth coordinate $E\!-\!X\!Y\!Z$. (b): The time cost for different modules under different point number of $Pcl_s$. The average value is marked in the corresponding color.}
      \label{fig12}
      \vspace{-0.3cm}
   \end{figure}

 We set up a hardware test environment as Fig. \ref{fig11}(b), the drone takes off behind the boxes and then a person enters the FOV of the camera and walks straight after the drone passes static boxes to test the effectiveness of our method. In Fig. \ref{fig11}(d), the camera is fixed and takes photos every 0.7 s during the flight. Seven photos are composed together. In the 4th frame the walking person appeared, the green line shows the trajectory of the drone while the blue dash line is for the person. It can be concluded that the reaction of the drone is similar to the simulation above. The data for this hardware test is shown in Fig. \ref{fig12}, the velocity curve indicates that the drone reacts very fast once the moving obstacle appears. To analyze the change of time cost , $Pcl_s$ only use the latest data frame so its size varies. In Fig. \ref{fig12}(b), the blue line is only for line 1 in \textbf{Algorithm \ref{alg5}} (static collision check) and the red line represents the whole \textbf{Algorithm \ref{alg5}}. The blue line has a strong positive linear correlation with time, because the number of the input points of the collision check procedure determines the distance calculation times. While the red and yellow line shows the irregularity, because the moving obstacle brings external computation burden to \textbf{Algorithm \ref{alg5}}, and the number of moving obstacles has no relation with the point cloud size. 

At last, we compare our work with SOTA in Table \ref{table_3}. Since all related works differ significantly from ours in terms of application background and test platform, for numeral indicators we only compare the time cost for reference. The abbreviations stand for: obs (obstacle), atc (average time cost). ``Optimality'' refers to if the trajectory planner considers the minimal acceleration or the shortest trajectory to avoid the moving obstacle. ``N/A'' refers to the work that gets obstacle information from an external source and \textbf{does not include} obstacle detection procedure. Most works have severe restrictions on the obstacle type, and do not consider the optimality. Our work has a great advantage in generality and optimality, the time cost is also the shortest unless the event camera is considered.

\vspace{-0.3cm}
   
\vspace{-0.3cm}
\begin{table}[h]
\caption{comparison between different works}
\vspace{-0.7cm}
\label{table_3}
\begin{center}
\begin{tabular}{|c|c|c|c|c|}
\hline
Work & Sensor type & Obs type limits & Time cost &Optimality \\
\hline
\cite{c12} & Sonar & dynamic obs & 1-2 s & No\\
\hline
\cite{c3} & N/A & dynamic obs & 0.045-0.13 s& No\\
\hline
\cite{c22} & Camera\&Lidar & \textbf{No} & About 0.1 s  & No\\
\hline
\cite{c2}& N/A & human & 0.2-0.3 s & No\\
\hline
\cite{c14} & Event camera & dynamic obs & \textbf{3.5 ms} & No
\\
\hline
\textbf{Ours}& Depth camera & \textbf{No} & \textbf{Atc$\leq$0.02 s} & \textbf{Yes}\\
\hline
\end{tabular}
\end{center}
\vspace{-0.5cm}
\end{table}

\vspace{-0.12cm}
\section{CONCLUSION AND FUTURE WORK}

In this paper, we present a \textbf{computationally efficient} algorithm framework for \textbf{both static and dynamic} obstacle avoidance for UAVs based \textbf{only on point cloud}. The test results indicate our work is feasible and show great promise in practical applications.

However, when the speed or angular velocity of drone is high, and also because of narrow FOV of single-camera, recognizing moving obstacles becomes significantly difficult and unreliable. In future research, we intend to improve the robustness of our method in aggressive flights and test it with different sensors such as lidar.


\addtolength{\textheight}{-6cm}   






\begin{thebibliography}{29}

\bibitem{c18} J. Chen, T. Liu, and S. Shen, “Online generation of collision-free trajectories for quadrotor flight in unknown cluttered environments,” in 2016 IEEE International Conference on Robotics and Automation (ICRA),vol. 2016, 2016, pp. 1476–1483.
\bibitem{c24} B. Damas and J. Santos-Victor, “Avoiding moving obstacles: the forbidden velocity map,” in 2009 IEEE/RSJ International Conference on Intelligent Robots and Systems, 2009, pp. 4393–4398.
\bibitem{c26} H. Chen and P. Lu, “Computationally efficient obstacle avoidance trajectory planner for uavs based on heuristic angular search method,” IROS 2020, 2020. arXiv:2003.06136.
\bibitem{c20} C. Yu, Z. Liu, X.-J. Liu, F. Xie, Y. Yang, Q. Wei, and Q. Fei, “Ds-slam: A semantic visual slam towards dynamic environments,” in 2018IEEE/RSJ International Conference on Intelligent Robots and Systems(IROS), 2018, pp. 1168–1174.
\bibitem{c21} J. Kim and Y. Do, “Moving obstacle avoidance of a mobile robot using a single camera,” Procedia Engineering, vol. 41, pp. 911–916, 2012.
\bibitem{c2} T. Nageli, J. Alonso-Mora, A. Domahidi, D. Rus, and O. Hilliges, “Real-time motion planning for aerial videography with real-time with dynamic obstacle avoidance and viewpoint optimization,” in IEEE Robotics and Automation Letters, vol. 2, no. 3, 2017, pp. 1696–1703.
\bibitem{c23} A. Ess, B. Leibe, K. Schindler, and L. van Gool, “Moving obstacle detection in highly dynamic scenes,” in 2009 IEEE International Conference on Robotics and Automation, 2009, pp. 4451–4458.
\bibitem{c22} A. Cherubini, F. Spindler, and F. Chaumette, “Autonomous visual navigation and laser-based moving obstacle avoidance,” IEEE Transactions on Intelligent Transportation Systems, vol. 15, no. 5, pp. 2101–2110, 2014.
\bibitem{c14} D. Falanga, K. Kleber, and D. Scaramuzza, “Dynamic obstacle avoidance for quadrotors with event cameras.” Science Robotics, vol. 5,no. 40, 2020.
\bibitem{c3} N. Malone, H.-T. Chiang, K. Lesser, M. Oishi, and L. Tapia, “Hybrid dynamic moving obstacle avoidance using a stochastic reachable set-based potential field,” IEEE Transactions on Robotics, vol. 33, no. 5,pp. 1124–1138, 2017.
\bibitem{c15} H. Febbo, J. Liu, P. Jayakumar, J. L. Stein, and T. Ersal, “Moving obstacle avoidance for large, high-speed autonomous ground vehicles,” in 2017 American Control Conference (ACC), 2017, pp. 5568–5573.
\bibitem{c7} W. Luo, W. Sun, and A. Kapoor, “Multi-robot collision avoidance under uncertainty with probabilistic safety barrier certificates,” in Advances in Neural Information Processing Systems, vol. 33, 2020.
\bibitem{c12} W. Zhang, S. Wei, Y. Teng, J. Zhang, X. Wang, and Z. Yan, “Dynamic obstacle avoidance for unmanned underwater vehicles based on an improved velocity obstacle method.” Sensors, vol. 17, no. 12, p. 2742, 2017.
\bibitem{c4} S. Liu, M. Watterson, S. Tang, and V. Kumar, “High speed navigation for quadrotors with limited on board sensing,” in 2016 IEEE International Conference on Robotics and Automation (ICRA), 2016, pp.1484–1491.
\bibitem{c5} H. Chen, P. Lu, and C. Xiao, “Dynamic obstacle avoidance for uavs using a fast trajectory planning approach,” in 2019 IEEE International Conference on Robotics and Biomimetics (ROBIO), 2019, pp. 1459–1464.
\bibitem{c6} M. Watterson and V. Kumar, “Safe receding horizon control for aggressive mav flight with limited range sensing,” in 2015 IEEE/RSJ International Conference on Intelligent Robots and Systems (IROS), 2015,pp. 3235–3240.

\bibitem{c8} H. Oleynikova, Z. Taylor, R. Siegwart, and J. Nieto, “Safe local exploration for replanning in cluttered unknown environments for microaerial vehicles,” in IEEE Robotics and Automation Letters, vol. 3, no. 3, 2018,pp. 1474–1481.
\bibitem{c10} T. M. Howard, C. J. Green, A. Kelly, and D. Ferguson, “State space sampling of feasible motions for high-performance mobile robot navigation in complex environments,” in Journal of Field Robotics - Special Issue on Field and Service Robotics archive, vol. 25, no. 6, 2008, pp.325–345.
\bibitem{c11} B. Zhou, F. Gao, L. Wang, C. Liu, and S. Shen, “Robust and efficient quadrotor trajectory generation for fast autonomous flight,” in IEEE Robotics and Automation Letters, vol. 4, no. 4, 2019, pp. 3529–3536.
\bibitem{c25} E.-H. Guechi, J. Lauber, and M. Dambrine, “On-line moving-obstacle avoidance using piecewise bezier curves with unknown obstacle trajectory,” in 2008 16th Mediterranean Conference on Control and Automation, 2008, pp. 505–510.
\bibitem{c13} J. Bialkowski, M. Otte, S. Karaman, and E. Frazzoli, “Efficient collision checking in sampling-based motion planning via safety certificates,” The International Journal of Robotics Research, vol. 35, no. 7, pp. 767–796,2016.
\bibitem{c9} M. W. Mueller, M. Hehn, and R. DAndrea, “A computationally efficient motion primitive for quadrocopter trajectory generation,” IEEE Transactions on Robotics, vol. 31, no. 6, pp. 1294–1310, 2015.
\bibitem{c27} M. Ester, H.-P. Kriegel, J. Sander, and X. Xu, “A density-based algorithm for discovering clusters in large spatial databases with noise,” in Proc. 1996 Int. Conf. Knowledg Discovery and Data Mining (KDD’96), 1996, pp. 226–231.













\end{thebibliography}
\end{document}